\documentclass[conference, a4paper, 10.5pt]{IEEEtran}
\IEEEoverridecommandlockouts

\usepackage{cite}
\usepackage{amsmath,amssymb,amsfonts}
\usepackage{algorithmic}
\usepackage{graphicx}
\usepackage{textcomp}
\usepackage{xcolor}
\usepackage{subfigure}
\usepackage{multirow}
\usepackage{comment}

\usepackage[left=1.62cm,right=1.62cm,top=1.9cm, bottom=3.56cm]{geometry}
\usepackage{setspace}
\def\BibTeX{{\rm B\kern-.05em{\sc i\kern-.025em b}\kern-.08em
    T\kern-.1667em\lower.7ex\hbox{E}\kern-.125emX}}
\begin{document}

\title{Adversarial Detector with Robust Classifier}

\author{\IEEEauthorblockN{1\textsuperscript{st} Takayuki Osakabe}
\IEEEauthorblockA{\textit{Tokyo Metropolitan University} \\
Tokyo, Japan \\
osakabe-takayuki@ed.tmu.ac.jp}
\and
\IEEEauthorblockN{2\textsuperscript{nd} Maungmaung Aprilpyone}
\IEEEauthorblockA{\textit{Tokyo Metropolitan University} \\
Tokyo, Japan \\
april-pyone-maung-maung@ed.tmu.ac.jp}
\and
\IEEEauthorblockN{3\textsuperscript{rd} Sayaka Shiota}
\IEEEauthorblockA{\textit{Tokyo Metropolitan University} \\
Tokyo, Japan \\
sayaka@tmu.ac.jp}
\and
\IEEEauthorblockN{4\textsuperscript{th} Hitoshi Kiya}
\IEEEauthorblockA{\textit{Tokyo Metropolitan University} \\
Tokyo, Japan \\
kiya@tmu.ac.jp}
}

\maketitle

\begin{abstract}
Deep neural network (DNN) models are well-known to 
easily misclassify prediction results by using input
images with small perturbations, called 
adversarial examples. 
In this paper, we propose a novel adversarial detector, 
which consists of a robust classifier and a plain one, 
to highly detect adversarial examples. 
The proposed adversarial detector is carried out in 
accordance with the logits of plain and robust classifiers. 
In an experiment, the proposed detector is demonstrated 
to outperform a state-of-the-art detector without any robust classifier.
\end{abstract}


\section{Introduction}
Deep neural networks (DNNs) have been widely employed 
in many fields such as computer vision. 
In particular, image classification is a very important task 
as an application of DNNs. 
However, DNN models are well-known to easily misclassify
prediction results due to the use of adversarial examples 
that are input images including 
small perturbations \cite{AdvWarning, AdvTraining}. 
Because of the problem with DNN models, many countermeasures
have been studied so far. Countermeasures 
against adversarial examples are classified into two approaches. 
One is to robustly train DNN models against 
adversarial examples \cite{AdvTraining, Kiya, maung_access, AdvTraining2, AdvTraining3, AdvTraining4, SHF, maung_ICIP}. 
The other is to detect adversarial examples prior 
to a classifier \cite{Mahalanobis, ML-LOO, IF-Detect, filter-Detect, AdvDetect, AdvDetect2}.

In this paper, we focus on the latter approach. 
The proposed novel adversarial detector 
consists of a robust classifier and a plain one, 
and it is carried out by using 
the logits of the two classifiers. 
In an experiment, the proposed detector is demonstrated 
to outperform a state-of-the-art detector 
under some conditions.

\section{Related work}
\subsection{Adversarial attacks}
\begin{figure}[tb]
\centering
\includegraphics[width=8.5cm]{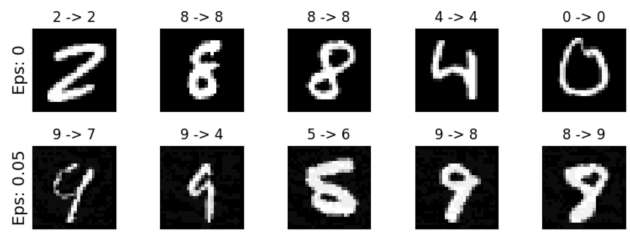}
\caption{Clean (1st row) and adversarial examples (2nd row)}
\label{adv_exp}
\end{figure}

Adversarial attacks are a malicious attack in which 
an attacker intentionally creates data to cause 
misclassification in a classifier.  
Adversarial examples are created by
adding a small noise to the input data.
An example of adversarial examples is shown 
in Fig. \ref{adv_exp}.
As shown in Fig. \ref{adv_exp},
there is no way to distinguish 
between clean and adversarial samples, 
but misclassification is caused.

Adversarial attacks can be classified into 
non-target attacks and target attacks.
In non-target attacks, an attacker tries to 
make input data misclassify so that
it is far away from the original class of the input data.
In contrast, in target attacks, an attacker tries to mislead 
input data to a specified target class.
In this paper, we mainly focus on target attacks.
In this section, we summarize four adversarial attack methods 
considered in this paper: 
fast gradient sign method (FGSM) \cite{AdvTraining},
projected gradient descent (PGD) \cite{PGD}, 
Jacobian-based saliency map attack (JSMA) \cite{JSMA}, 
and Carlini and Wagner attack (CW) \cite{CW}.

\begin{figure*}[t]
\centering
\includegraphics[width=17.5cm]{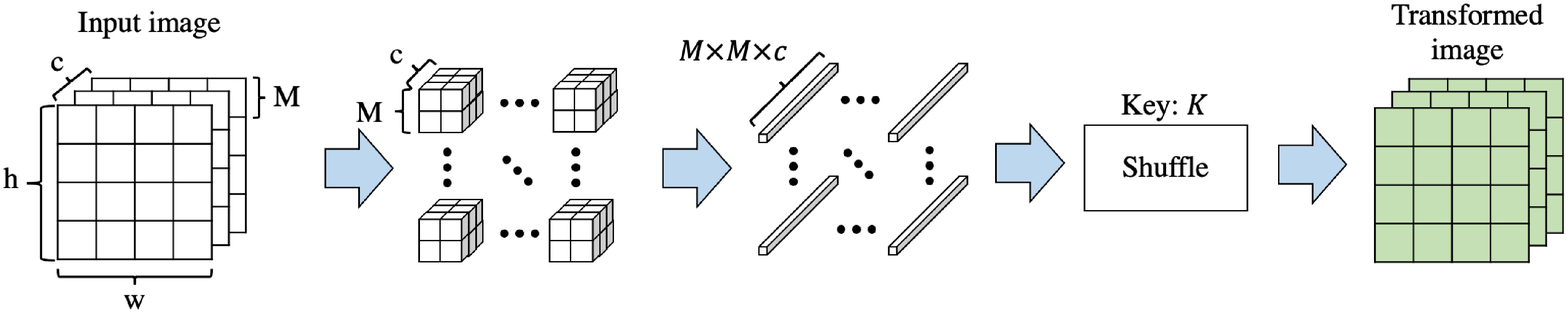}
\caption{Procedure of pixel shuffling}
\label{SHF}
\end{figure*}

\textbf{FGSM}: This is one of the simplest and fastest adversarial attack methods.
An attacker linearly fits the cross entropy loss around 
a target sample, and perturbs input image pixels as 
maximizing a gradient loss in one-step. 
FGSM is explained as
\begin{align}
\begin{split}
	x_{adv} = x + \epsilon \cdot sign( \nabla_x J(\theta, x, y))
\label{FGSM}
\end{split}
\end{align}
where $\nabla_x J$ is the gradient of a loss function
with respect to an original input $x$,
$y$ is the ground truth label of $x$,
$\epsilon$ is a perturbation added to $x$,
and $\theta$ represents classification model parameters.

\textbf{PGD}: PGD is an attack method, which is 
an extension of FGSM.
In FGSM, perturbation $\epsilon$ is added to 
input $x$ in a single step,
while input $x$ is gradually changed with 
step size $\alpha$ in PGD.
The pixel values of a perturbed image are clipped 
so that they do not change more than $\pm{\epsilon}$
from the original pixel value.
PGD attack is shown in the following equation.
\begin{align}
	x_{k+1} = Clip_{(x+\epsilon,x-\epsilon)}	(x_k+\alpha *sign(\nabla x_k))
\label{PGD}
\end{align}

\textbf{JSMA}: This method is iterative and costly.
This attack uses $L_0$ norm to attack 
one or two pixels which cause the largest change in the loss.
An input image is attacked by adjusting parameters $\theta$, 
which represents the magnitude of a perturbation applied to 
each target pixel, and $\gamma$, which controls
the percentage of pixels to be perturbed.

\textbf{CW}: This attack creates an adversarial example by 
searching for the smallest perturbation computed in $L_0$, 
$L_1$, and $L_2$ norms.
This attack is carried out by controlling parameter $C$ called as confidence.
If we set a high value of this parameter,
an adversarial example is more different from an original input.

\section{Proposed detector}
\begin{figure}[t]
\hspace{0.8cm}
\includegraphics[width=8cm]{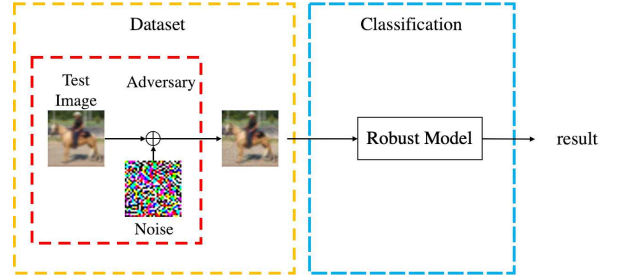}
\caption{Image classification system with robust classifier}
\label{robust}
\end{figure}

There are two approaches for defending models against adversarial examples. 
The first approach is to design a classifier that is 
robust against adversarial attacks as shown in Fig. \ref{robust} \cite{AdvTraining, AdvTraining2, AdvTraining3, AdvTraining4, SHF}. 
This approach includes methods for training models with
a dataset including adversarial examples \cite{AdvTraining}, 
and training models with images transformed with a secret key \cite{SHF}. 
The proposed detector includes a robust classifier \cite{AdvTraining}. 
Three image transformation methods were used for 
the robust classifier: pixel shuffling, bit flipping, and 
format-preserving, feistel-based encryption (FFX). 
We use pixel shuffling as image transformation to train a
robust classifier. 
Pixel shuffling is carried out in the following steps (see Fig. \ref{SHF}).
\begin{enumerate}
	\item Split an input image with a size of $h \times w \times c$ into blocks with a size of $M \times M$.
	\item Flatten each block into a vector with a length of $M \times M \times c$.
	\item Shuffling elements in each vector with a common key $K$ to each block.
	\item Merge the transformed blocks.
\end{enumerate}

\begin{figure}[t]
\hspace{0.2cm}
\includegraphics[width=9cm]{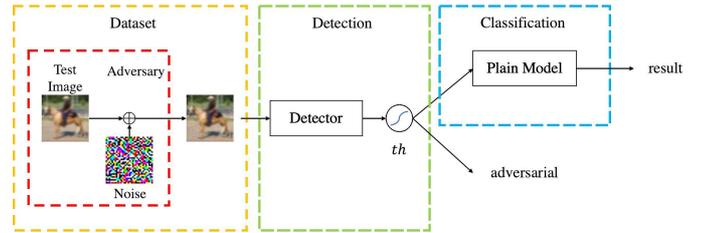}
\caption{Image classification system with adversarial example detector}
\label{detect}
\end{figure}

The other approach is to detect adversarial examples just before a classifier as shown in Fig. \ref{detect} \cite{Mahalanobis, ML-LOO, IF-Detect, filter-Detect, AdvDetect, AdvDetect2}.

\begin{figure}[t]
	\begin{center}
	\subfigure[Clean samples]{
		\includegraphics[width=4.1cm]{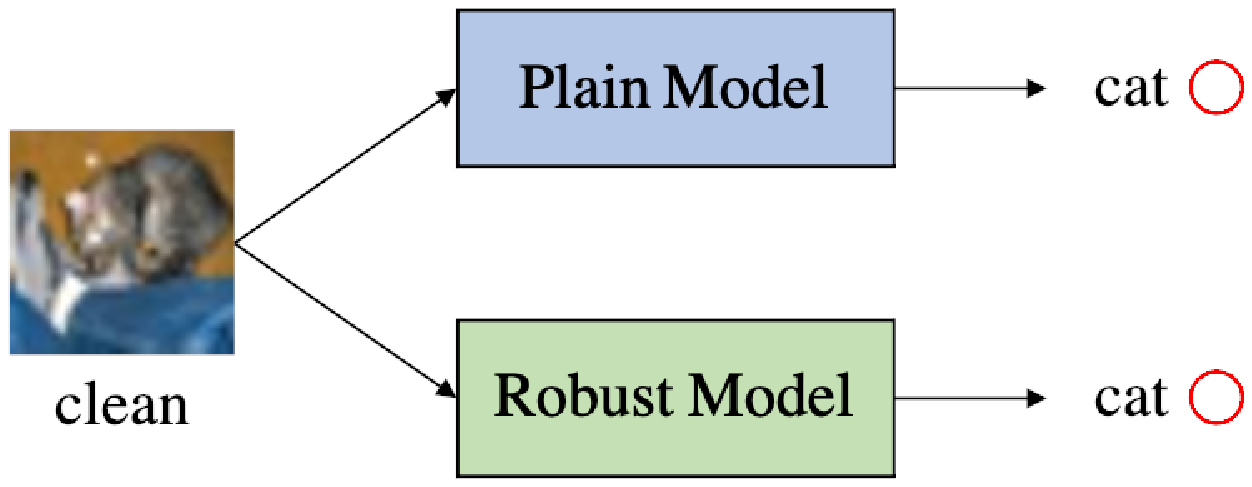}}
	\subfigure[Adversarial samples]{
		\includegraphics[width=4.1cm]{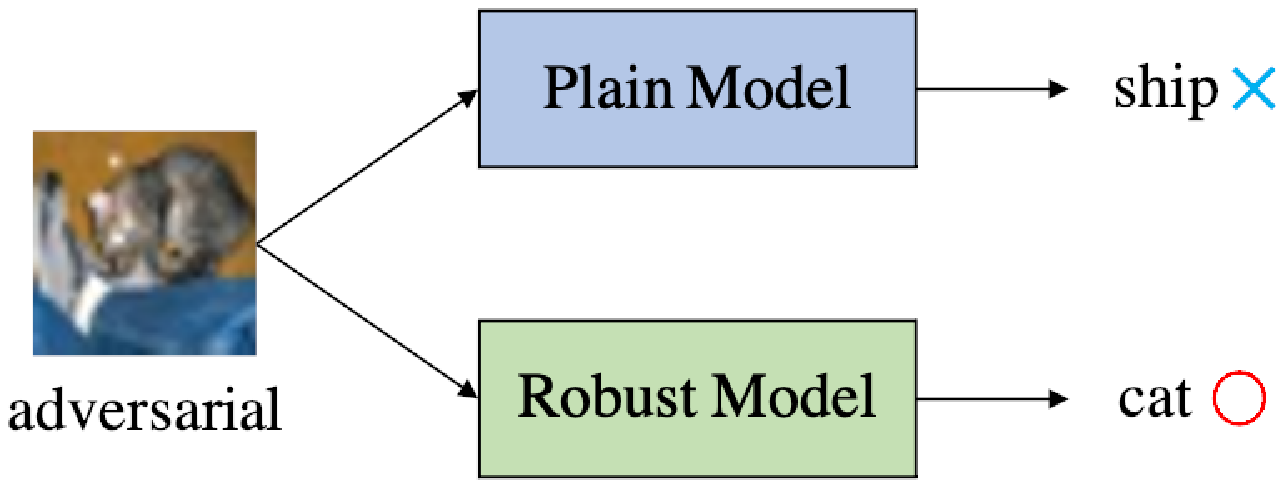}}
	\end{center}
	\caption{Assumptions in proposed method}
	\label{explain_detector}
\end{figure}

\begin{figure}[tb]
\centering
\includegraphics[width=8.5cm]{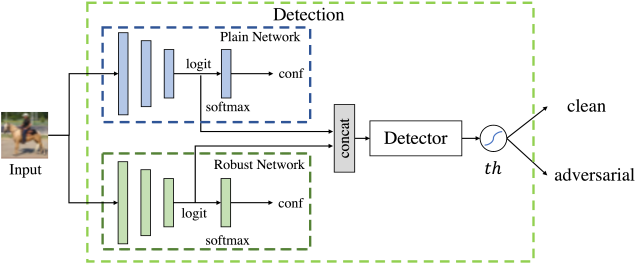}
\caption{Proposed adversarial detector}
\label{proposed_detector}
\end{figure}

In this paper, we focus on methods for detecting adversarial examples,
and propose a novel detection method, which consists 
of plain and robust classifiers.
In the proposed method, it is expected that there is 
a difference between the output of a plain classifier and 
that of a robust one if the input image is an adversarial example, 
as shown in Fig. \ref{explain_detector}. 
In other words, the output of a plain classifier is expected 
to be the same as that of a robust classifier if the input
image is clean. 
The final output from the softmax layer in a classifier, i.e. a
confidence value, is represented as a positive value 
in the range [0,1] for each label. 
Furthermore, the sum of all confidence values from each classifier is 1.
To relax   these  constraints, in this paper, 
two logits obtained from the plain
classifier and robust classifier are concatenated, 
and they are used to decide whether an input image is 
an adversarial example or not, instead of confidence
values. The above procedure is summarized in Fig. \ref{proposed_detector}.

\section{Experiments}
In the experiment, the effectiveness of 
the proposed detector was evaluated on 
the basis of two metrics:  accuracy (Acc) and 
area under the curve (AUC), given by Eqs. \eqref{eq:acc} to \eqref{eq:auc}.

\begin{align}
	Acc &= \frac{TP+TN}{TP+FP+TN+FN} \label{eq:acc} \\
	TPR &= \frac{TP}{TP+FN} \label{eq:tpr} \\
	FPR &= \frac{FP}{TN+FP} \label{eq:fpr} \\
	AUC &= \int TPR\,d(FPR) \label{eq:auc}
\end{align}

These metrics are based on the confusion matrix in binary classification
shown in Table \ref{Binary}.

\begin{table}[t]
  \begin{center}
    \caption{Confusion matrix in binary classification}
    \begin{tabular}{c|c|c|c|}
    	\multicolumn{2}{c}{} & \multicolumn{2}{c}{Predicted} \\ \cline{3-4}
        \multicolumn{2}{c|}{} & Positive & Negative \\ \cline{2-4}
      	\multirow{2}{*}{\rotatebox{90}{Actual}} & Positive  & True Positive & False Negative \\\cline{2-4}
      		& Negative  &  False Positive   &  True Negative  \\ \cline{2-4}
    \end{tabular}
    \label{Binary}
  \end{center}
\end{table}

\subsection{Experimental setup}
We used the CIFAR-10 dataset for testing and training detectors, 
where the dataset consists of 60,000 images
(50,000 images for training, 10,000 images for testing).
In the experiment, we assume a white-box attack, and four attacks:
FGSM \cite{AdvTraining}, 
PGD \cite{PGD}, 
CW \cite{CW} and 
JSMA \cite{JSMA}, 
were applied to input images. 
We set parameters for each of these attacks as $\epsilon=8/255$ 
for FGSM and PGD,
confidence parameter $C=0$ for CW, 
and $\theta=1.0, \gamma=0.1$ for JSMA. 
8,000 clean images from the test set, 
and 8,000 adversarial examples generated from 
the clean images were  used to train detectors.  
The other 2,000 clean images and 2,000 adversarial examples 
generated from them were used to test a detector.
The effectiveness of the proposed detector was compared 
with Lee’s method \cite{Mahalanobis}. 
ResNet-18 \cite{resnet} was used for both a plain classifier 
and a robust one for the proposed method, 
and the robust classifier was trained 
in accordance with 
Maung’s method \cite{SHF}.

Table \ref{cls_performance} shows the classification performance of 
the plain and robust classifiers on 
10,000 test images under various attacks to show 
the robustness of the classifiers.

\begin{table}[t]
  \caption{Classification performance of plain and robust classifiers (Noise parameters; $\epsilon=8/255$ for FGSM and PGD. Confidence parameter $C=0$ for CW. $\gamma=0.1$ and $\theta=1.0$ for JSMA)}
  \label{cls_performance}
  \centering
  \begin{tabular}{cccccc}
    \hline
    \multirow{2}{*}{Classifier}  & \multicolumn{5}{c}{Attack} \\ 
      & CLEAN & FGSM & PGD & CW & JSMA \\ \hline
     Plain  & 0.952 & 0.557 & 0.100 & 0.100 & 0.104 \\
     Robust & 0.916 & 0.812 & 0.882 & 0.911 & 0.743 \\ \hline
  \end{tabular}
\end{table}

\subsection{Experimental results}
In the experiment, the same attacks were used for testing
and training detectors.

AUC and Acc scores are shown 
in Tables \ref{detector_auc} and \ref{detector_acc}, respectively.
From the tables, Lee’s method outperformed
the proposed detector under FGSM,
but the proposed method outperformed Lee’s method
under the other attacks.
The reason is that FGSM is not a strong attack as described in 
Table \ref{cls_performance}, so it is difficult to detect adversarial
examples from a difference between the plain and robust classifiers.
Adversarial detection methods are required to maintain 
a high detection accuracy under strong attacks, 
since weak attacks do not give serious damages in general.

To evaluate how well our detection method can be transferred 
to unseen attacks, we trained detectors on the features 
obtained using the CW attack with $C=0$,
and then evaluated them on the other (unseen) attacks.
Experimental results against the unseen attacks
are shown in Table \ref{unseen}.
It can be observed that our proposed method showed the best
performance except to FGSM.
Our detection method is based on the output
of a plain classifier and that of a robust one,
so the proposed detecor is transferable under the condition of
using the attacks.

\begin{table}[t]
  \caption{AUC of proposed and Lee's detectors}
  \label{detector_auc}
  \centering
  \begin{tabular}{ccccc}
    \hline
    \multirow{2}{*}{Detector}  & \multicolumn{4}{c}{Attack} \\ 
      & FGSM & PGD & CW & JSMA \\ \hline
     Lee  & \textbf{0.994} & 0.983 & 0.727 & 0.921 \\
     Proposed  & 0.805 & \textbf{1.000} & \textbf{0.952} & \textbf{0.952} \\ \hline
  \end{tabular}
\end{table}

\begin{table}[t]
  \caption{Acc of proposed and Lee's detectors}
  \label{detector_acc}
  \centering
  \begin{tabular}{ccccc}
    \hline
    \multirow{2}{*}{Detector}  & \multicolumn{4}{c}{Attack} \\ 
      & FGSM & PGD & CW & JSMA \\ \hline
     Lee  & \textbf{0.990} & 0.974 & 0.598 & 0.865 \\
     Proposed  & 0.740 & \textbf{0.999} & \textbf{0.939} & \textbf{0.942} \\ \hline
  \end{tabular}
\end{table}

\begin{table}[t]
  \caption{Acc performance for unseen attacks. Detectors trained with adversarial examples generated by CW attack. (Noise parameters; $\epsilon=8/255$ for FGSM and PGD. Confidence parameter $c=0$ for CW. $\gamma=0.1$ and $\theta=1.0$ for JSMA)}
  \label{unseen}
  \centering
  \scalebox{0.8}{
  \begin{tabular}{ccccc}
    \hline
    \multirow{2}{*}{Detector}  & \multicolumn{4}{c}{Attack} \\ 
      & CW(seen) & FGSM(unseen) & PGD(unseen) & JSMA(unseen) \\ \hline
     Lee  & - & \textbf{0.971} & 0.960 & 0.695 \\
     Proposed  & - & 0.553 & \textbf{0.961} & \textbf{0.940} \\ \hline
  \end{tabular}}
\end{table}

\section{Conclusions}
In this paper, we propose a detection method for
adversarial examples that consists of two image classifiers.
In the experiment, the proposed method was confirmed
to be able to maintain a high accuracy even under
the use of strong attacks.
We also showed that our proposed method is robust against 
unseen attacks under the limited condition.

\section*{acknouledgement}
This study was partially supported by JSPS KAKENHI (Grant Number JP21H01327)

\bibliographystyle{IEEEtran}
\bibliography{papers}

\begin{thebibliography}{10}
\providecommand{\url}[1]{#1}
\csname url@samestyle\endcsname
\providecommand{\newblock}{\relax}
\providecommand{\bibinfo}[2]{#2}
\providecommand{\BIBentrySTDinterwordspacing}{\spaceskip=0pt\relax}
\providecommand{\BIBentryALTinterwordstretchfactor}{4}
\providecommand{\BIBentryALTinterwordspacing}{\spaceskip=\fontdimen2\font plus
\BIBentryALTinterwordstretchfactor\fontdimen3\font minus
  \fontdimen4\font\relax}
\providecommand{\BIBforeignlanguage}[2]{{%
\expandafter\ifx\csname l@#1\endcsname\relax
\typeout{** WARNING: IEEEtran.bst: No hyphenation pattern has been}%
\typeout{** loaded for the language `#1'. Using the pattern for}%
\typeout{** the default language instead.}%
\else
\language=\csname l@#1\endcsname
\fi
#2}}
\providecommand{\BIBdecl}{\relax}
\BIBdecl

\bibitem{AdvWarning}
C.~Szegedy, W.~Zaremba, I.~Sutskever, J.~Bruna, D.~Erhan, I.~Goodfellow, and
  R.~Fergus, ``Intriguing properties of neural networks,'' in
  \emph{International Conference on Learning Representations (ICLR)}, Apr.
  2014.

\bibitem{AdvTraining}
I.~Goodfellow, J.~Shlens, and C.~Szegedy, ``Explaining and harnessing
  adversarial examples,'' in \emph{International Conference on Learning
  Representations (ICLR)}, May. 2015.

\bibitem{Kiya}
\BIBentryALTinterwordspacing
H.~Kiya, M.~AprilPyone, Y.~Kinoshita, I.~Shoko, and S.~Shiota, ``An overview of
  compressible and learnable image transformation with secret key and its
  applications,'' in \emph{arXiv:2201.11006}, 2022. [Online]. Available:
  \url{https://arxiv.org/abs/2201.11006}
\BIBentrySTDinterwordspacing

\bibitem{maung_access}
M.~AprilPyone, Y.~Kinoshita, and H.~Kiya, ``Adversarial robustness by one bit
  double quantization for visual classification,'' \emph{IEEE Access}, vol.~7,
  pp. 177\,932--177\,943, 2019.

\bibitem{AdvTraining2}
A.~Kurakin, I.~J. Goodfellow, and S.~Bengio, ``Adversarial machine learning at
  scale,'' in \emph{International Conference on Learning Representations
  (ICLR)}, Apr. 2017.

\bibitem{AdvTraining3}
T.~Miyato, S.~ichi Maeda, M.~Koyama, K.~Nakae, and S.~Ishii, ``Distributional
  smoothing with virtual adversarial training,'' in \emph{International
  Conference on Learning Representations (ICLR)}, May. 2015.

\bibitem{AdvTraining4}
F.~Tram`er, A.~Kurakin, N.~Papernot, I.~Goodfellow, D.~Boneh, and P.~McDaniel,
  ``Ensemble adversarial training: Attacks and defenses,'' in
  \emph{International Conference on Learning Representations (ICLR)}, May.
  2018.

\bibitem{SHF}
M.~AprilPyone and H.~Kiya, ``Block-wise image transformation with secret key
  for adversarially robust defense,'' \emph{IEEE Trans. on Information
  Forensics and Security}, vol.~16, pp. 2709--2723, Mar. 2021.

\bibitem{maung_ICIP}
M.~Aprilpyone and H.~Kiya, ``Encryption inspired adversarial defense for visual
  classification,'' in \emph{IEEE International Conference on Image Processing
  (ICIP)}, Oct. 2020, pp. 1681--1685.

\bibitem{Mahalanobis}
K.~Lee, K.~Lee, H.~Lee, and J.~Shin, ``A simple unified framework for detecting
  out-of-distribution samples and adversarial attacks,'' in \emph{Advances in
  Neural Information Processing Systems (NIPS)}, vol.~31, Dec. 2018.

\bibitem{ML-LOO}
P.~Yang, J.~Chen, C.-J. Hsieh, J.-L. Wang, and M.~I. Jordan, ``Ml-loo:
  Detecting adversarial examples with feature attribution,'' in
  \emph{Association for the Advancement of Artificial Intelligence (AAAI)},
  vol.~34, Feb. 2020, pp. 6639--6647.

\bibitem{IF-Detect}
G.~Cohen, G.~Sapiro, and R.~Giryes, ``Detecting adversarial samples using
  influence functions and nearest neighbors,'' in \emph{IEEE Conference on
  Computer Vision and Pattern Recognition (CVPR)}, Jun. 2020, pp.
  14\,453--14\,462.

\bibitem{filter-Detect}
A.~Higashi, M.~Kuribayashi, N.~Funabiki, H.~H. Nguyen, and I.~Echizen,
  ``Detection of adversarial examples based on sensitivities to noise removal
  filter,'' in \emph{Asia Pacific Signal and Information Processing Association
  (APSIPA)}, Dec. 2020, pp. 1386--1391.

\bibitem{AdvDetect}
X.~Li and F.~Li, ``Adversarial examples detection in deep networks with
  convolutional filter statistics,'' in \emph{International Conference on
  Computer Vision (ICCV)}, Oct. 2017, pp. 5764--5772.

\bibitem{AdvDetect2}
J.~H. Metzen, T.~Genewein, V.~Fischer, and B.~Bischoff, ``On detecting
  adversarial perturbations,'' in \emph{International Conference on Learning
  Representations (ICLR)}, Apr. 2017.

\bibitem{PGD}
A.~Madry, A.~Makelov, L.~Schmidt, D.~Tsipras, and A.~Vladu, ``Towards deep
  learning models resistant to adversarial attacks,'' in \emph{International
  Conference on Learning Representations (ICLR)}, May. 2018.

\bibitem{JSMA}
N.~Papernot, P.~D. McDaniel, S.~Jha, M.~Fredrikson, Z.~B. Celik, and A.~Swami,
  ``The limitations of deep learning in adversarial settings,'' in \emph{IEEE
  European Symposium on Security and Privacy (EuroS\&P)}, Mar. 2016.

\bibitem{CW}
N.~Carlini and D.~A. Wagner, ``Towards evaluating the robustness of neural
  networks,'' in \emph{IEEE Symposium on Security and Privacy (SP)}, May. 2017,
  pp. 39--57.

\bibitem{resnet}
K.~He, X.~Zhang, S.~Ren, and J.~Sun, ``Deep residual learning for image
  recognition,'' in \emph{IEEE Conference on Computer Vision and Pattern
  Recognition (CVPR)}, Jun. 2016, pp. 770--778.

\end{thebibliography}

\end{document}